# The Comparison of Methods Artificial Neural Network with Linear Regression Using Specific Variables for Prediction Stock Price in Tehran Stock Exchange

Reza Gharoie Ahangar,
Master of Business Administration
of Islamic Azad University – Babol
branch &Membership of
young researcher club, Iran.

reza_gh578@yahoo.com

Mahmood Yahyazadehfar
Associate Professor of Finance
University of Mazandaran
Babolsar, Iran
m.yahyazadeh@umz.ac.ir

Hassan Pournaghshband
Professor of Computer Science
Department, Southern Polytechnic
State University

*Abstract*- **In this paper, researchers estimated the stock price of activated companies in Tehran (Iran) stock exchange. It is used Linear Regression and Artificial Neural Network methods and compared these two methods.**

**In Artificial Neural Network, of General Regression Neural Network method (GRNN) for architecture is used. In this paper, first, researchers considered 10 macro economic variables and 30 financial variables and then they obtained seven final variables including 3 macro economic variables and 4 financial variables to estimate the stock price using Independent components Analysis (ICA). So, we presented an equation for two methods and compared their results which shown that artificial neural network method is more efficient than linear regression method.**

## I. INTRODUCTION

The recent upsurge in research activities into artificial neural networks (ANNs) has proven that neural networks have powerful pattern classification and prediction capabilities. ANNs have been successfully used for a variety of tasks in many fields of business, industry, and science [24]

Interest in neural networks is evident from the growth in the number of papers published in journals of diverse scientific disciplines. A search of several major databases can easily result in hundreds or even thousands of "neural networks" articles published in one year.
A neural network is able to work parallel with input variables and consequently handle large sets of data quickly. The principal strength with the network is its ability to find patterns [3] ANNs provide a promising alternative tool for fore asters. The inherently nonlinear structure of neural networks is particularly useful for capturing the complex underlying relationship in many real world problems. Neural networks are perhaps more versatile methods for forecasting applications in that not only can they find nonlinear structures in a problem, they can also model linear processes. For example, the capability of neural networks in modeling linear time series has been studied and confirmed by a number of researchers [8],[11],[26].

One of the major application areas of ANNs is forecasting. There is an increasing interest in forecasting using ANNs in recent years. Forecasting has a long history and the importance of this old subject is reflected by the diversity of its applications in different disciplines ranging from business to engineering.
The ability to accurately predict the future is fundamental to many decision processes in planning, scheduling, purchasing, strategy formulation, policy making, and supply chain operations and stock price. As such, forecasting is an area where a lot of efforts have been invested in the past. Yet, it is still an important and active field of human activity at the present time and will continue to be in the future. A survey of research needs for forecasting has been provided by Armstrong [1].

Forecasting has been dominated by linear methods for many decades. Linear methods are easy to develop and implement and they are also relatively simple to understand and interpret. However, linear models have serious limitation in that they are not able to capture any nonlinear relationships in the data. The approximation of linear models to complicated nonlinear relationships is not always satisfactory. In the early 1980s, Makridakis (1982)organized a large-scale forecasting competition (often called M-competition) where a majority of commonly used linear methods were tested with more than 1,000 real time series. The mixed results show that no single linear model is globally the best, which may be interpreted as the failure of linear modeling in accounting for a varying degree of nonlinearity that is common in real world problems [10].
The financial forecasting or stock market prediction is one of the hottest fields of research lately due to its commercial applications owing to the high stakes and the kinds of attractive benefits that it has to offer [18]. Unfortunately, stock market is essentially dynamic, non-linear, complicated, nonparametric, and chaotic in nature [21]. The time series are multi-stationary, noisy, random, and has frequent structural breaks [13],[22]. In addition, stock market's movements are affected by many macro-economical factors ([Miao et al., 2007] and [Wang, 2003]) such as political events, firms' policies, general economic conditions, commodity price index, bank rate, bank exchange rate, investors' expectations, institutional investors' choices, movements of other stock market, psychology of investors, etc.[12],[22]

Artificial neural networks are one of the technologies that have made great progress in the study of the stock markets. Usually stock prices can be seen as a random time sequence with noise, artificial neural networks, as large-scale parallel processing nonlinear systems that depend on their own intrinsic link data, provide methods and techniques that can approximate any nonlinear continuous function, without a priori assumptions about the nature of the generating process, see (Pino et al., 2008).It is obvious that several factors are effective on future stock price and the main weak point in this surrey is that all of them considered a few limit factors in future stock price and using linear methods, Regarding that fact, although previous studies highlighted the problem to some extent, none of them didn't provide a comprehensive model to estimate the stock prince. If one estimates the prince and provides a model for it to eliminate uncertainties to a large extent it can help to increase the investments in stock exchange. Conducting the scientific surveys to obtain a suitable and desirable model to estimate the stock price is the best task. [16]

The purpose of this study is to insure the investors and provides them with suitable information for better investment. Regarding that future stock price and its estimation are important factors in accurate decision making for better investment for investors; therefore, this survey is aiming to estimate the future stock price of companies acting in Tehran (Iran) stock exchange using the most effective variables related to the stock price

## II. REVIEW OF LITERATURE

Eldon Y. Li(1994)

The purpose of his paper is to answer two of the most frequently asked questions: "What are neural networks?" "Why are they so popular in today's business fields?" The paper reviews the common characteristics of neural networks and discusses the feasibility of neural-net applications in business fields. It then presents four actual application cases and identifies the limitations of the current neural-net technology. [7]

Kyoung-jae Kimand & Ingoo Han(2000)

Their paper proposes genetic algorithms (GAs) approach to feature discrimination and the determination of connection weights for artificial neural networks (ANNs) to predict the stock price index. In this study, GA is employed not only to improve the learning algorithm, but also to reduce the complexity in feature space. GA optimizes simultaneously the connection weights between layers and the thresholds for feature discrimination. The genetically evolved weights mitigate the well-known limitations of the gradient descent algorithm. In addition, globally searched feature discretization reduces the dimensionality of the feature space and eliminates irrelevant factors. Experimental results show that GA approach to the feature discrimination model outperforms the other two conventional models. [9]

Shaikh A. Hamid and Zahid Iqbal(2003)

They present a primer for using neural networks for financial forecasting. They compare volatility forecasts from neural networks with implied volatility from S&P 500 Index futures options using the Barone-Adesi and Whaley (BAW) American futures options pricing model. Forecasts from neural networks outperform implied volatility forecasts and are not found to be significantly different from realized volatility. Implied volatility forecasts are found to be significantly different from realized volatility in two of three forecast horizons. [19]

D. E. Allen, W. Yang,(2004)

Examines the deviation of the UK total market index from market fundamentals implied by the simple dividend discount model and identifies other components that also affect price movements. The components are classified as permanent, temporary, excess stock return and non-fundamental innovations to stock prices by employing a multivariate moving-average model as applied in [J. Financial Quant. Anal. 33 (1998) 1] and imposing relevant restrictions on the model in light of Sims–Bernanke forecast error variance decomposition. They find that time-varying discounted rates play an active role in explaining price deviations. [5]

David Enke and Suraphan Thawornwong (2005)

Introduces an information gain technique used in machine learning for data mining to evaluate the predictive relationships of numerous financial and economic variables. Neural network models for level estimation and classification are then examined for their ability to provide an effective forecast of future values. A cross-validation technique is also employed to improve the generalization ability of several models. The results show that the trading strategies guided by the classification models generate higher risk-adjusted profits than the buy-and-hold strategy, as well as those guided by the level-estimation based forecasts of the neural network and linear regression models. [4]

Qing Cao, Karyl B. Leggio, Marc J. Schniederjans (2005)

Their study uses artificial neural networks to predict stock price movement (i.e., price returns) for firms traded on the Shanghai stock exchange. We compare the predictive power using linear models from financial forecasting literature to the predictive power of the univariate and multivariate neural network models. Our results show that neural networks outperform the linear models compared. These results are statistically significant across our sample firms, and indicate neural networks are a useful tool for stock price prediction in emerging markets, like China. [17]

Yi-Hsien Wang(2007)

This study integrated new hybrid asymmetric volatility approach into artificial neural networks option-pricing model to improve forecasting ability of derivative securities price. Owing to combines the new hybrid asymmetric volatility method can be reduced the stochastic and nonlinearity of the error term sequence and captured the asymmetric volatility simultaneously. Hence, in the ANNS option-pricing model, the results demonstrate that Grey-GJR–GARCH volatility provides higher predictability than other volatility approaches. [25]

Pei-Chann Chang andChen-Hao Liu(2008)

In their study, an integrated system, CBDWNN by combining dynamic time windows, case based reasoning (CBR), and neural network for stock trading prediction is developed and it includes three different stages: (1) screening out potential stocks and the important influential factors; (2) using back propagation network (BPN) to predict the buy/sell points (wave peak and wave trough) of stock price and (3) adopting case based dynamic window (CBDW) to further improve the forecasting results from BPN.

The empirical results show that the CBDW can assist the BPN to reduce the false alarm of buying or selling decisions. [15]

Sheng-Hsun Hsu and JJ Po-An Hsieh (2008)

Their study employs a two-stage architecture for better stock price prediction. Specifically, the self-organizing map (SOM) is first used to decompose the whole input space into regions where data points with similar statistical distributions are grouped together, so as to contain and capture the non-stationary property of financial series. After decomposing heterogeneous data points into several homogenous regions, support vector regression (SVR) is applied to forecast financial indices. The proposed technique is empirically tested using stock price series from seven major financial markets. The results show that the performance of stock price prediction can be significantly enhanced by using the two-stage architecture in comparison with a single SVR model. [20]

Wei-Sen Chen and Yin-Kuan Dua(2008)

Their paper adopted the operating rules of the Taiwan stock exchange corporation (TSEC) which were violated by those companies that were subsequently stopped and suspended, as the range of the analysis of this research. In addition, this paper also used financial ratios, other non-financial ratios, and factor analysis to extract adaptable variables. Moreover, the artificial neural network (ANN) and data mining (DM) techniques were used to construct the financial distress prediction model. The empirical experiment with a total of 37 ratios and 68 listed companies as the initial samples obtained a satisfactory result, which testifies for the feasibility and validity of their proposed methods for the financial distress prediction. [23]

Zhang Yudong and Wu Lenan(2008)

Their paper proposed an improved bacterial chemo taxis optimization (IBCO), which is then integrated into the back propagation (BP) artificial neural network to develop an efficient forecasting model for prediction of various stock indices. Experiments show its better performance than other methods in learning ability and generalization. [27]

E.L. de Faria and J.L. Gonzalez (2009)

Their work performs a predictive study of the principal index of the Brazilian stock market through artificial neural networks and the adaptive exponential smoothing method, respectively. The objective is to compare the forecasting performance of both methods on this market index, and in particular, to evaluate the accuracy of both methods to predict the sign of the market returns. Also the influence on the results of some parameters associated to both methods is studied. Their results show that both methods produce similar results regarding the prediction of the index returns. On the contrary, the neural networks outperform the adaptive exponential smoothing method in the forecasting of the market movement, with relative hit rates similar to the ones found in other developed markets. [6]

## III. OBJECTIVES

The present study attempts to undertake the following objectives:

1- Considering the main variables to estimate future stock price of companies acting in stock exchange.

2- Price estimation using two methods of artificial neural network and linear regression neural networks and comparison of these two methods' results.

## IV. RESEARCH METHODOLOGY

### A. Sample Unit

The population of present study including all companies who were acting in Tehran stock exchange in 1380– 1386. There fore, those companies whose symbol was not active during this period were omitted and finally, 100 companies were chosen. The scope of subject in this study includes the consideration of the relationship between macro economic and financial variables with stock future price. Scope of locative of this study includes all accepted companies who were active in Tehran stock exchange from early 1379 to the end of 1386.

### B. Data Collection Method

In this study, we used 10 macro economic variables and 30 financial variables to study their effects on stock future price. Data related to macro economic variables were collected through Central Bank yearbook, economic reports and balance sheet of Central Bank and Monetary and financial Research center of Iran Central Bank and data related to companies financial variables were calculated collected through companies financial statements sand informational Agency of Tehran(Iran) stock exchange.

## C. Methodology Steps

1. Identifying related factors and omitting additional variables (among macro economic and financial variables) through the analysis of independent components.
2. Modeling and estimating stock future price through the linear regression equation.
3. Modeling and estimating stock future efficiency using General regression neural network.
4. Comparison of result related to these methods.

### a) Independent Components Analysis (ICA)

To estimate financial time series, it is necessary to use a set of continuous descriptive input variables among a very huge set of primary inputs.

It is difficult to choose a significant and suitable subset of input variables. In several scientific fields, it is difficult to find a reasonable transfer for a huge set of multi-data.

Our purpose is to use a technique to summarize independent components of time series in a set of variables which is named independent components Analysis (ICA). This method will decrease the number of descriptive variables by decreasing a set of financial and economic information into smaller subsets of independent components and maintaining the suitable information. Removing the random elements from each data set will facilitate the identification of relationship between independent components and stock indexes.

Independent components Analysis are process to summarize a new set of statistical independent components in a guide vector. These components will show some estimations of data main resource.

This process supposes a matrix of time series which includes a compound process; so, this process will analyze the independent components by creating a matrix when we enter them, and identify the related and unrelated components and provide us with the best matrix of estimative variables.

This technique will summarize as follows:
• This technique will summarize independent components of time series in a set of variables.
• This technique will find a way to change data with the minimum statistical dependency among the summarized components into a linear data.
• If two random variables are unrelated, they will not be independent.
• This technique is so special for analysis and estimation which uses two matrixes of data covariance and data changes by increasing the arrangement of linear and non-linear regression.

### b) Linear Regression

If researcher wants to estimate the dependent variable by one or more independent variables, he will use a linear regression model. This model will be shown as follows.

Amount of P for each set of data will result in minimum $\mu$ .When ever we use standard scores instead of raw variables in the analysis, p regression coefficients will be shown as B. This relation will be shown as.

Linear regression can be a method to estimate a set of time series.

Average of financial and macro economic variables of identified resources in the beginning of each year are independent variables in these estimations. Dependent variables Q are the real output of the company in estimation model, which dependent on price data of all stocks in our sample. Dependent variable will be estimated using regression step method (OLS). All independent variables will enter to the regression equation. These in dependent variables with P values more than 5% will be omitted in estimation period and at last, we will choose a subset of independent variables. Olson & Mossman state that variables of 3 to 7 independent variable will show the best estimations for this period. According this study if step solution method chooses more than eight independent variables, P – value will be decreased to 3% or 4% , and if step solution method chooses one or two variables, P- value will be increased to 10% to include more variables.

$$Q_{j,t} = \sum_{i=1}^{k} P_{i,t} * F_{j,i,t-1} + u_{j,t} \qquad (1)$$

K = the number of independent variables
P = regression coefficient of independent variable I in month t
$F_{j,i,t-1}$ = independent variable I for stock j at the end of previous period (month t-1).
$Uj,t$ = error terms for each regression
$Qj,t$ = price of (dependent variable) stock j in month t

### c) General Regression Neural Network

GRNN can approximate any arbitrary function from historical data. The major strength of GRNN compared to other Ann's is that its internal structure is not problem dependent.

Topology of GRNN
• GRNN consists of four layers:
• The first layer is responsible for reception of information.
• The input neurons present the data to the second layer (pattern neurons).
• The output of the pattern neurons are forwarded to the third layer (summation neurons).
• summation neurons are sent to the fourth layer (output neuron)

And we can summarize this model as:
• This model will consider a few non- linear aspects of the estimation problem.
• This network model will be taught immediately, and will be suitable for scattered data.

• First, data will be clustered to decrease the needed layers in hidden layer.
• This model enables to solve any problems in monotonous functions.
• This model can not ignore non-related inputs with out the main revisions in the main algorithm.

## V. FACTORS FOR COMPARISION OF TWO METHODS RESULTS

In time series, it is very important to conform an estimation model to data pattern and we can obtain the conformity of estimation method with data pattern by calculating estimation error during the time period.

For example, when a technique of estimation estimates the periodical and seasonal alternations in time series, then estimation error will show the disordered or random component in time series.

Error square mean index is obtained through dividing total error differences square by time series. Error percent absolute value mean is an index which will be used whenever estimation of error based on percent is more suitable. Determination coefficient is the most important factor one can explain the relationship between two variants by which.

1- MSE
2- MAPE
3- $R^2$

## VI. CHOOSING FINAL VARIABLES AMONG PRIMARY VARIABLES

40 financial and macroeconomic variables will enter independent components analysis method:

### A. Macroeconomic Variables
Growth rates of industrial production
Inflation rate
Interest rate
Exchange rate
Rate of return on stock public
Unemployment rate
Oil price
Gross Domestic product (GDP)
Money supply 1 (M1)
Money supply 2 (M2)

### B. Financial Variables
Book value per share
Sales per share
Earning per share
Cash flow per share
Inventory turnover rate
Annual average volume of daily trading relative to annual average total market capitalization
Dividend yield
Dividend payout ratio
Dividend per share
Total of sales to total assets
Bid – ask spread
Market impact of a trade
Price per share
Trading volume
Turnover rate
Commission rate
Indicator variables for the day of the week effect
Holiday effect
January month
Amortized effective spread
Price history
Past return
Size of firm
Ratio of total debt to stockholder's equity
Pastor measure
Ratio of absolute stock return to dollar volume
Market depth
Ratio of net income to book equity
Operating income to total assets
Operating income to total sales

Independent components analysis (ICA) most method chooses variables with minimum statistical dependency and explanation strength, and then we chose 40 variables

### C. Financial Variables
Earning per share
Size of firm
Ratio of total debt to stockholder's equity
Operating income to total sales

### D. Macroeconomic Variables
Inflation rate
Money supply 1 (M1)
Growth rates of industrial production

## VII. RESULTS AND ANALYSES

Here, we show the results of two methods and the model created by linear regression and neural network methods and comparison of the models' results using the above-mentioned factors.

## A. Estimation of Linear Regression Model

Table I: Model Summary

| Durbin-Watson | Std. Error of The Estimate | Adjusted R Square | R Square | R | Model |
|---|---|---|---|---|---|
| 2,013 | 83,487569 | 0,211 | 0,279 | 0,368 | 1 |

A  Predictors: (Constant), EXCHANGE, DEPT, EPS, SOF, INFLATION, M1
B Dependent Variable: Stock Price

Table II: table of ANOVA

| Model | | Sum Squares | df | Mean Square | F | sig |
|---|---|---|---|---|---|---|
| 1 | Regression | 381258,653 | 7 | 441,257 | 5,009 | 0,000a |
| | Residual | 287923,471 | 1117 | | | |
| | Total | 326214,368 | 1123 | 382,186 | | |

a.  Predictors: (Constant), EPS, SOF, income, inflation, M1, Dept, ratio

Table III: table of Coefficients a

| Model | Unstanddardized Coefficients | | Standardized Coefficients | t | sig |
|---|---|---|---|---|---|
| | B | Std. Error | Beta | | |
| 1 (Constant) | -14,61 | 39,216 | | -2,498 | 0,459 |
| ratio | 2,009 | 0,843 | 2,138 | 3,181 | 0,001 |
| inflation | 7,162 | 3,728 | 0,179 | 2,772 | 0,005 |
| income | -0.208 | 0.096 | -0,022 | -0,532 | 0,066 |
| Dept | 0.0309 | 0,223 | 0,031 | 1,991 | 0,042 |
| SOF | -0,0001 | 0,001 | 0,027 | 2,107 | 0,047 |
| EPS | 0,189 | 0,005 | 0,184 | 2,987 | 0,001 |

a.   Dependent Variable: stock price

$$Y=-14.61+2.009X_1+7.162X_2+0.0309X_3-0.0001X_4+0.189X_5 \qquad (2)$$

Y: stock price
$X_1$: Growth rate of industrial products
$X_2$: Inflation rate
$X_3$: Ratio of total liabilities to stockholders pay
$X_4$: Company's degree
$X_5$: Earning per share

As it is observed financial variable of operational income to total selling are not mentioned in the model should be more and than 1.98 and Sig less than 0.05, respectively, for a variable to be meaningful and mentioned in the model. There fore, the significance level for this variable is more than 5% and t value is (-0.532), so this variable will not be mentioned in the model.

According the tables which calculated by algebra method, multi correlation factor (R) is 0.368. That is, it is 0.368 correlations between independent variables and dependent variables. This means that independent variables which remained in regression equation are 0.368 and have a significant relation ship with stock price.

Coefficient of determination ($R^2$) or (Pearson's correlation coefficient) show a ratio of total dependent variable changes which are calculated by dependent variables of the equation. So, in dependent variables could estimate 0.279 variance of dependent variable (price). Moreover, according the B standard coefficient one can say growth rate variable of industrial products (Beta = 2.138) in a significant level 0.001 is the most descriptive for dependent variable value or stock price.

### B. Estimation of General Regression Neural Network Model

To estimate General Regression Neural Network Model, we consider 7 variables obtained from dependent components analysis as input (P) and stock price as output (T).

Also, we calculated spread = 0.8326.becaues spread of more than 1 will cause in hyper fitting of network and a larger region of input to output vector. And its very small value will cause in increase of estimation error. In a way that function will have a high slope and neuron which weights are more similar to its input will have more outputs than other neurons. In this network member of input vector P will be calculated for all neurons and will be calculated for transfer function (sigmoid function) and the output will be gained after multiplying in weights vector and adding to bias. And this output will be a vector. We used a 3- layer general regression neural network which had seven neurons in internal layer and fourteen neurons in middle layer and one neuron in external to design.

After using learning algorithm and network education of 37 educational periods, network error graph is as follows.

### a) Model One

Is an estimated model which is not educated and has its own real error?

$$Y = -8.11 + 1.83X_1 - 0.000011X_2 + 7.16X_3 + 2.07X_4 - .00008X_5 + 0.957X_6 + 0.243X_7 \qquad (3)$$

### b) Model Two

- Which is obtained through using learning algorithm in model One which has the minimum error.

- LM learning algorithm was chosen which has the most adaptability to all survey aspects.

- Value of SPREAD = 0.8326 was used because spread value of more than 1 will case in hype fitting in network and

a larger region of input to output vector, and its very small value will cause in increase of estimation error.

- We used a 3 – layer general regression neural network which had seven neurons in internal layer, fourteen neurons in middle layer and one neuron in external layer to design.

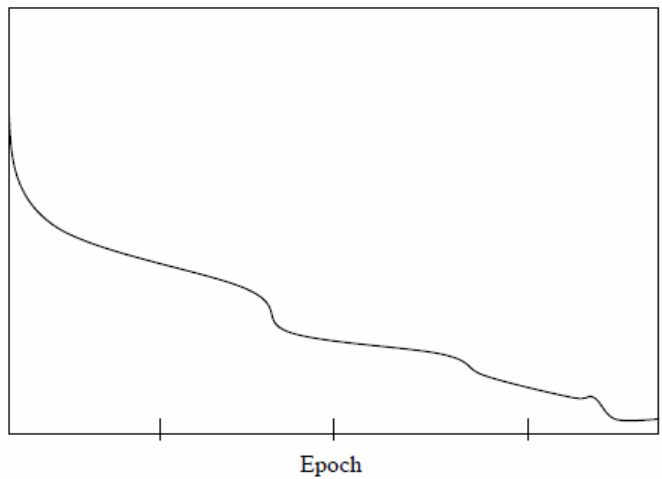

Fig1: Mean Squared Error of GRNN Network

$$Y = -11.07 + 4.11X1 - 0.000009X2 + 6.74X3 + 1.31X4 - 0.0007X5 + 0.39X6 + 0.131X7 \qquad (4)$$

## VIII. CONCLUSION

### A. Comparison of two Methods Results

As it is shown in table below, value of estimation error square mean, absolute mean of error percent and ($R^2$)

coefficient will be decreased significantly after using training in neural network which will be shown the increase of estimation factor in trained neural network.

Table IV: the compare of two methods

| $R^2$ | MAPE | MSE | |
|---|---|---|---|
| 0,71 | 1,42 | 76,2 | GRNN |
| 0,368 | 3,73 | 97,6 | regression linear |

After using LM algorithm and network training, above statistics will be changed as follows

Table V: the compare of two methods after using LM algorithm

| R2 | MAPE | MSE | |
|---|---|---|---|
| 0,98 | 0,78 | 31,6 | GRNN |

Neural networks are quicker than other methods including regression because they are executing parallel and tolerate more errors and also these networks can make rules without any implicit formula which are understandable in an environment of chaos and implicitly, such as stock exchange which is a very important factor.

As said before, in this survey, we chose 100 companies of high quality in Tehran stock exchange. And we understand that artificial neural network method is better than linear regression method in estimation. And neural network method is more descriptive than linear regression method, and at last, the research theory, neural network method is better than linear regression method, is approved in this study.